\documentclass[conference]{IEEEtran}
\IEEEoverridecommandlockouts

\usepackage{cite}
\usepackage{amsmath,amssymb,amsfonts}

\usepackage{algorithm}
\usepackage{algpseudocode}

\usepackage{graphicx}
\usepackage{textcomp}
\usepackage{xcolor}
\def\BibTeX{{\rm B\kern-.05em{\sc i\kern-.025em b}\kern-.08em
    T\kern-.1667em\lower.7ex\hbox{E}\kern-.125emX}}
\begin{document}

\title{Enhancing Adversarial Robustness of Deep Neural Networks Through Supervised Contrastive Learning\\

}

\author{
\IEEEauthorblockN{
Longwei Wang,
Navid Nayyem,
Abdullah Rakin
}

\IEEEauthorblockA{Department of Computer Science, University of South Dakota\\
}

}

\maketitle

\begin{abstract}
Adversarial attacks exploit the vulnerabilities of convolutional neural networks by introducing imperceptible perturbations that lead to misclassifications, exposing weaknesses in feature representations and decision boundaries. This paper presents a novel framework combining supervised contrastive learning and margin-based contrastive loss to enhance adversarial robustness. Supervised contrastive learning improves the structure of the feature space by clustering embeddings of samples within the same class and separating those from different classes. Margin-based contrastive loss, inspired by support vector machines, enforces explicit constraints to create robust decision boundaries with well-defined margins. Experiments on the CIFAR-100 dataset with a ResNet-18 backbone demonstrate robustness performance improvements in adversarial accuracy under Fast Gradient Sign Method attacks.
\end{abstract}


\section{Introduction}

Convolutional Neural Networks have revolutionized the field of computer vision by achieving remarkable performance on tasks such as image classification, object detection, semantic segmentation, and more. Their ability to learn hierarchical feature representations from data has been instrumental in solving complex problems that were once considered intractable. However, alongside their success, CNNs have been found to exhibit a critical vulnerability: they are highly sensitive to adversarial attacks. Adversarial attacks are small, imperceptible perturbations added to input data that can lead to drastic changes in model predictions. For instance, a correctly classified image of a bird can be misclassified as an airplane with the addition of a carefully crafted perturbation. This phenomenon raises serious concerns about the reliability and robustness of CNNs, particularly in real-world, safety-critical applications such as autonomous driving, medical imaging, and security systems.

A primary cause of this adversarial vulnerability lies in the feature extraction process of CNNs. During training, the convolutional kernels in CNNs often fail to learn robust and meaningful features. Instead, they tend to rely on spurious patterns or shortcuts present in the training data. These fragile patterns can easily be manipulated by adversarial perturbations, causing the model to misclassify inputs. This inadequacy stems from the standard training objective, which typically optimizes for task-specific accuracy (e.g., cross-entropy loss) without explicitly encouraging the model to learn invariant and robust feature representations. As a result, the learned kernels are insufficiently generalized and prone to failure under even minor deviations from the training distribution.

Numerous approaches have been proposed to improve the adversarial robustness of CNNs. These include:
\begin{itemize}
    \item \textbf{Adversarial Training}: Incorporating adversarial examples during training to make the model robust against specific types of attacks. While effective, this method is computationally expensive and often fails to generalize to unseen attacks.
    \item \textbf{Regularization Techniques}: Methods such as weight decay, dropout, and input noise injection aim to improve generalization. However, these are often insufficient to counteract strong adversarial perturbations.
    \item \textbf{Defensive Distillation}: A technique that modifies the model training process to reduce sensitivity to small input changes. While promising, it can be circumvented by advanced attack strategies.
    \item \textbf{Gradient Regularization}: Penalizing large gradients in the loss function with respect to input can reduce sensitivity to perturbations but may also hinder model performance on clean data.
\end{itemize}

Despite their contributions, these methods share common shortcomings. Most are tailored to specific types of adversarial attacks, making them less effective against new or unseen threats. Furthermore, they often fail to address the underlying issue: the extraction of robust and meaningful features by the CNN kernels.

To address these challenges, we propose a novel approach leveraging \textbf{Supervised Contrastive Learning} to improve the adversarial robustness of CNNs. Contrastive learning is a self-supervised learning paradigm that aims to learn feature representations by comparing similar (positive) and dissimilar (negative) samples. Supervised Contrastive Learning extends this framework by incorporating label information, allowing the model to learn class-specific relationships in the feature space. Specifically, it encourages samples from the same class to cluster tightly together while maintaining clear separation from samples of other classes. This explicit alignment of features makes the model's decision boundaries more robust to adversarial perturbations.

Our approach combines the supervised contrastive loss with the traditional cross-entropy loss to achieve two objectives simultaneously:
\begin{itemize}
    \item \textbf{Robust Feature Learning}: The supervised contrastive loss aligns feature representations of the same class in the latent space, making them less susceptible to adversarial manipulations.
    \item \textbf{Task-Specific Accuracy}: The cross-entropy loss ensures that the model retains its ability to perform well on the primary classification task.
\end{itemize}




This paper makes the following key contributions:
\begin{itemize}
    \item \textbf{Identification of Kernel Limitations}: We analyze the limitations of CNN kernels in extracting robust features and establish this as a primary cause of adversarial vulnerability.
    \item \textbf{Supervised Contrastive Learning Framework}: We introduce a novel training framework that combines supervised contrastive learning with task-specific cross-entropy loss to improve feature robustness.
    \item \textbf{Experimental Validation}: Extensive experiments on CIFAR-100 demonstrate that our approach improves adversarial robustness under FGSM attacks, outperforming the baseline model.
\end{itemize}

\subsection{Structure of the Paper}
The rest of the paper is organized as follows:
\begin{itemize}
    \item \textbf{Section 2}: Reviews related work on adversarial robustness and contrastive learning.
    \item \textbf{Section 3}: Introduces the supervised contrastive learning framework and the combined loss function.
    \item \textbf{Section 4}: Details the experimental setup and presents the results, including evaluations under adversarial attacks.
    \item \textbf{Section 5}: Concludes the paper and discusses potential directions for future research.
\end{itemize}

\section{Related Work}

\subsection{Adversarial Robustness in CNNs}
The vulnerability of Convolutional Neural Networks (CNNs) to adversarial attacks has been a critical area of research since Szegedy et al. (2014) revealed that neural networks exhibit unexpected fragility under adversarial perturbations \cite{szegedy2014intriguing}. These perturbations, often imperceptible to humans, exploit the complex and non-linear decision boundaries of CNNs, resulting in misclassification of inputs.

\textbf{Adversarial Attacks:} Goodfellow et al. (2015) introduced the Fast Gradient Sign Method (FGSM) as one of the earliest efficient methods to generate adversarial examples \cite{goodfellow2015explaining}. This was followed by stronger iterative attacks like Projected Gradient Descent (PGD) \cite{madry2018towards} and optimization-based methods such as the Carlini and Wagner (CW) attack \cite{carlini2017towards}. These attacks highlighted how easily CNNs could be deceived, even under minimal perturbations.

\textbf{Adversarial Defenses:} To counter adversarial attacks, researchers have explored multiple defense strategies:
\begin{itemize}
    \item \textbf{Adversarial Training:} This approach, formalized by Madry et al. (2018), involves training models on adversarially perturbed examples, effectively hardening them against specific attacks \cite{madry2018towards}. While adversarial training is widely regarded as one of the most robust defenses, it incurs significant computational overhead and often struggles to generalize across unseen attacks \cite{tramer2018ensemble}.
    \item \textbf{Gradient Regularization:} Ross and Doshi-Velez (2018) proposed input gradient regularization as a way to smooth the model’s decision boundaries and reduce sensitivity to input perturbations \cite{ross2018improving}. However, this method is limited in handling high-dimensional perturbations.
    \item \textbf{Certified Robustness:} Approaches such as randomized smoothing \cite{cohen2019certified} and interval bound propagation (IBP) \cite{gowal2018effectiveness} provide theoretical guarantees of robustness under certain perturbation bounds. Despite their theoretical appeal, these methods often struggle with scalability and general applicability.
    \item \textbf{Feature Space Regularization:} Zhang et al. (2019) introduced TRADES, which explicitly balances robustness and accuracy by penalizing feature space discrepancies during adversarial training \cite{zhang2019theoretically}. Such methods have gained attention for providing a principled way to address robustness without overly compromising clean accuracy.
\end{itemize}

While these methods represent significant progress, they often rely on either computationally expensive procedures or attack-specific adaptations. Furthermore, many fail to address the root cause of adversarial vulnerability: the inadequacy of CNN kernels in extracting robust, meaningful features that generalize well beyond clean data.

\subsection{Contrastive Learning}
Contrastive learning has emerged as a cornerstone of self-supervised learning, offering a mechanism to learn high-quality representations without labeled data. The contrastive paradigm works by maximizing agreement between positive pairs (augmented views of the same sample) and minimizing agreement with negative pairs (different samples). SimCLR \cite{chen2020simple} and MoCo \cite{he2020momentum} have been pivotal in demonstrating the efficacy of contrastive learning for unsupervised representation learning.

\textbf{Supervised Contrastive Learning:} Khosla et al. (2020) extended the contrastive framework to the supervised setting by incorporating label information to create positive and negative pairs. In supervised contrastive learning, embeddings of samples belonging to the same class are brought closer in the latent space, while embeddings of different classes are pushed apart \cite{khosla2020supervised}. This additional supervision significantly improves intra-class clustering and inter-class separation, making the learned representations more discriminative and robust.

\textbf{Contrastive Learning for Adversarial Robustness:} While contrastive learning has primarily been explored for unsupervised and semi-supervised learning, its potential for improving adversarial robustness is gaining attention. Jiang et al. (2020) proposed adversarial contrastive learning, which combines contrastive objectives with adversarial training \cite{jiang2020robust}. However, the lack of label supervision in their approach limits the discriminative power of the learned features. Similarly, work by Fan et al. (2021) explored self-supervised adversarial learning but struggled to match the robustness of supervised methods \cite{fan2021defending}.

In this work, we leverage the strengths of supervised contrastive learning to enhance the robustness of CNN kernels. By encouraging class-specific feature alignment, we hypothesize that the learned representations become inherently resistant to adversarial perturbations.

\subsection{Combined Loss Frameworks}
Combining loss functions to achieve complementary objectives has been a widely studied approach in deep learning. In adversarial contexts, methods such as MART \cite{wang2019improving} and TRADES \cite{zhang2019theoretically} introduced auxiliary objectives to balance robustness and accuracy. These frameworks typically penalize adversarial discrepancies while optimizing for clean data performance.

In contrastive settings, loss combinations have proven effective in semi-supervised and unsupervised learning. MixMatch \cite{berthelot2019mixmatch} and FixMatch \cite{sohn2020fixmatch} introduced contrastive-like consistency regularization to improve label efficiency. Similarly, frameworks like SupCon \cite{khosla2020supervised} demonstrated the utility of combining supervised contrastive loss with standard classification objectives.

Xie et al. (2020) explored a hybrid approach that combined adversarial training with natural data, achieving improvements in both clean and robust accuracy \cite{xie2020adversarial}. However, their method relied heavily on computationally intensive adversarial generation. By contrast, our framework integrates supervised contrastive learning with cross-entropy loss, enabling efficient and scalable robust representation learning without the need for adversarial training.

Our proposed combined loss framework aligns robust feature learning with task-specific objectives, enabling CNN kernels to extract meaningful, class-specific features while enhancing resilience to adversarial perturbations.

\section{Methodology}

\subsection{Motivation for Supervised Contrastive Learning}

Adversarial attacks exploit weaknesses in convolutional neural networks (CNNs) by introducing imperceptible perturbations that lead to incorrect predictions. These vulnerabilities arise primarily due to:
(1) Weak feature representations, where CNNs rely on non-generalizable patterns susceptible to adversarial perturbations, and (2) poorly defined decision boundaries, allowing minor perturbations to push inputs across these boundaries.

Existing defenses, such as adversarial training, address these issues but suffer from high computational cost and limited generalization to unseen attack types. In contrast, Supervised Contrastive Learning (SCL) provides a scalable and generalizable solution by structuring the feature space through alignment of embeddings. By clustering embeddings of samples from the same class and separating embeddings from different classes, SCL creates robust and well-separated decision boundaries. These structured embeddings enhance resilience to adversarial perturbations while maintaining computational efficiency, motivating its use for refining adversarial robustness.

\subsection{Supervised Contrastive Learning for Robustness}

\begin{figure*}[t]
    \centering
    \includegraphics[width=\textwidth, keepaspectratio]{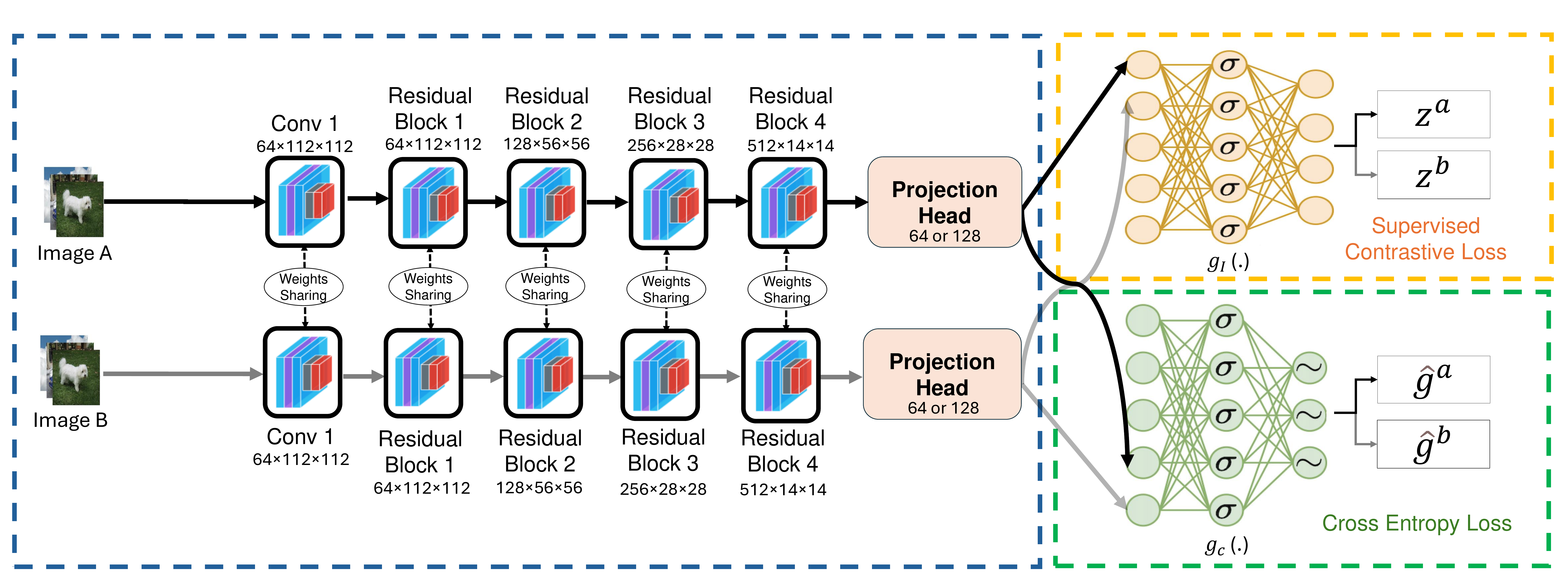}
    \caption{Supervised Contrastive Learning for enhanced robustness: neural network architecture with shared weights, residual blocks, projection heads, and loss functions. The architecture consists of two parallel streams processing input images A and B, with shared weights across residual blocks. Each stream outputs two projection vectors, which are used for computing Cross Entropy Loss and Supervised Contrastive Loss.}
    \label{fig:architecture}
\end{figure*}

Supervised Contrastive Learning leverages label information to guide the structuring of the feature space. For a batch of samples \(\{x_i\}_{i=1}^N\) with corresponding labels \(\{y_i\}_{i=1}^N\), the \textit{Supervised Contrastive Loss} is defined as:
\begin{equation}
\mathcal{L}_{SCL} = \frac{1}{N} \sum_{i=1}^N \frac{-1}{|P(i)|} \sum_{p \in P(i)} \log \frac{\exp(\text{sim}(\mathbf{z}_i, \mathbf{z}_p) / \tau)}{\sum_{a \in A(i)} \exp(\text{sim}(\mathbf{z}_i, \mathbf{z}_a) / \tau)},
\end{equation}
where \(N\) is the batch size, \(\mathbf{z}_i = f_\text{proj}(f_\text{backbone}(x_i))\) represents the normalized embedding of the \(i\)-th sample obtained from the projection head applied to the backbone features, \(P(i)\) contains positive samples (samples with the same label as \(x_i\)), and \(A(i)\) includes all samples in the batch except \(x_i\). The cosine similarity \(\text{sim}(\mathbf{z}_i, \mathbf{z}_j)\) is computed as:
\begin{equation}
\text{sim}(\mathbf{z}_i, \mathbf{z}_j) = \frac{\mathbf{z}_i \cdot \mathbf{z}_j}{\|\mathbf{z}_i\| \|\mathbf{z}_j\|}.
\end{equation}

The temperature hyperparameter \(\tau\) controls the sharpness of the similarity distribution. By minimizing \(\mathcal{L}_{SCL}\), the model aligns embeddings of positive samples while separating embeddings of negative samples, resulting in a structured feature space.

To refine a pre-trained baseline model, the supervised contrastive loss is combined with the cross-entropy loss:
\begin{equation}
\mathcal{L}_{\text{refined}} = \alpha \mathcal{L}_{SCL} + \beta \mathcal{L}_{CE},
\end{equation}
where \(\mathcal{L}_{CE}\) is the cross-entropy loss, and \(\alpha\) and \(\beta\) are weighting factors that balance the contributions of the contrastive and task-specific objectives. This combination ensures robust feature learning and strong classification performance.

\subsection{Margin-Based Contrastive Loss}

Supervised contrastive learning lacks explicit mechanisms to enforce well-defined boundaries between classes. To address this limitation, we propose the \textit{Margin-Based Contrastive Loss}, inspired by the margin maximization principle in Support Vector Machines (SVMs). This loss introduces explicit constraints to enhance intra-class compactness and inter-class separation.

The margin-based contrastive loss is defined as:
\begin{equation}
\begin{split}
\mathcal{L}_{\text{Margin}} = \frac{1}{N} \sum_{i=1}^N \Bigg[ &\frac{1}{|P(i)|} \sum_{p \in P(i)} \max(0, m_p - \text{sim}(\mathbf{z}_i, \mathbf{z}_p)) \\
&+ \frac{1}{|N(i)|} \sum_{n \in N(i)} \max(0, \text{sim}(\mathbf{z}_i, \mathbf{z}_n) - m_n) \Bigg].
\end{split}
\end{equation}
where \(P(i)\) and \(N(i)\) represent the sets of positive and negative samples for anchor \(i\), respectively, and \(m_p > 0\) and \(m_n > 0\) are the positive and negative margins.

The first term penalizes positive pairs with similarity scores below \(m_p\), ensuring tight clustering within classes. The second term penalizes negative pairs with similarity scores above \(m_n\), enforcing larger gaps between classes. These constraints lead to:
\begin{itemize}
    \item \textbf{Enhanced Feature Compactness}: Positive pairs form tighter intra-class clusters.
    \item \textbf{Improved Class Separation}: Negative pairs are pushed farther apart, creating robust decision boundaries.
    \item \textbf{Adversarial Resilience}: Larger margins make decision boundaries more stable against adversarial perturbations.
\end{itemize}

The total loss function for training the refined model combines the margin-based contrastive loss with the cross-entropy loss:
\begin{equation}
\mathcal{L}_{\text{total}} = \alpha \mathcal{L}_{\text{Margin}} + \beta \mathcal{L}_{CE},
\end{equation}
where \(\alpha\) and \(\beta\) control the relative contributions of the two objectives. This formulation ensures robust and discriminative feature learning while preserving classification performance.

Margin-based contrastive learning builds upon supervised contrastive learning by explicitly enforcing intra-class compactness and inter-class separation through well-defined margin constraints. These enhancements result in a feature space that is more robust to adversarial perturbations and better suited for high-stakes applications requiring reliable decision-making.

\subsection{Architecture}
The architecture is designed for improving robustness by Supervised Contrastive Learning. It consists of two parallel streams that process input images through convolutional layers and residual blocks with shared weights. These streams output projection vectors for computing classification and contrastive losses, enabling robust learning. Figure~\ref{fig:architecture} illustrates the detailed architecture.

\section{Experiments}

This section provides a comprehensive evaluation of the proposed framework, which leverages supervised contrastive learning (SCL) and margin-based contrastive loss to enhance adversarial robustness. The evaluation includes training a baseline ResNet-18 model, refining it using SCL and margin-based constraints, and assessing its robustness under adversarial attacks. Detailed analyses of results and ablation studies are presented to highlight the contributions of each component.

\subsection{Experimental Setup}

The experiments are conducted on the CIFAR-100 dataset, a widely used benchmark for image classification tasks, consisting of 60,000 images across 100 classes, with 50,000 training samples and 10,000 test samples. Each image has a resolution of $32 \times 32$ pixels, and the dataset is split into non-overlapping training and test sets.

\textbf{Data Preprocessing:} The dataset is normalized to the range $[-1, 1]$ using the mean and standard deviation of the CIFAR-100 dataset. No data augmentation techniques are applied to maintain a focus on the proposed robustness enhancements.

\textbf{Network Architecture:} ResNet-18 is employed as the baseline model due to its simplicity and effectiveness. The final fully connected layer is modified to output 100 logits, corresponding to the 100 classes. For the refined model, a two-layer projection head is added, producing 64-dimensional embeddings suitable for contrastive learning.

\textbf{Training Configurations:} The experiments are conducted in two stages:

1. \textit{Model Training:} The baseline ResNet-18 model is trained for 10 epochs using cross-entropy loss with a batch size of 128. The Adam optimizer is used with a learning rate of 0.001.

2. \textit{Model Refinement:} The trained baseline model is refined for 10 additional epochs using supervised contrastive learning combined with cross-entropy loss. The weighting factors for the combined loss are set to $\alpha = 0.5$ and $\beta = 0.5$.

\textbf{Margin-Based Refinement:} For experiments incorporating margin-based constraints, the margin-based contrastive loss is used with positive and negative margins set to $m_p = 0.5$ and $m_n = 0.1$, respectively. These margins are chosen based on hyperparameter tuning to achieve the best trade-off between intra-class compactness and inter-class separation.

\subsection{Evaluation Metrics}

To assess the performance of the proposed framework, the following metrics are used: Classification accuracy under adversarial attacks, including:
   \begin{itemize}
       \item \textit{Fast Gradient Sign Method (FGSM):} A single-step attack defined as:
       \begin{equation}
       \mathbf{x}_{adv} = \mathbf{x} + \epsilon \cdot \text{sign}(\nabla_{\mathbf{x}} \mathcal{L}),
       \end{equation}
       where $\epsilon = 0.01$ is the perturbation magnitude.

   \end{itemize}

\subsection{Model training loss:}

To evaluate the effectiveness of the proposed framework, we trained eight distinct models on the CIFAR-100 dataset, examining various configurations to highlight the impact of data augmentation, supervised contrastive learning (SCL), and margin-based constraints. The loss dynamics of each model were analyzed over 200 epochs, providing detailed insights into the behavior of baseline and refined models under different training regimes.

\subsubsection{Baseline Models}

\begin{figure}[t]
    \centering
    \includegraphics[width=\linewidth]{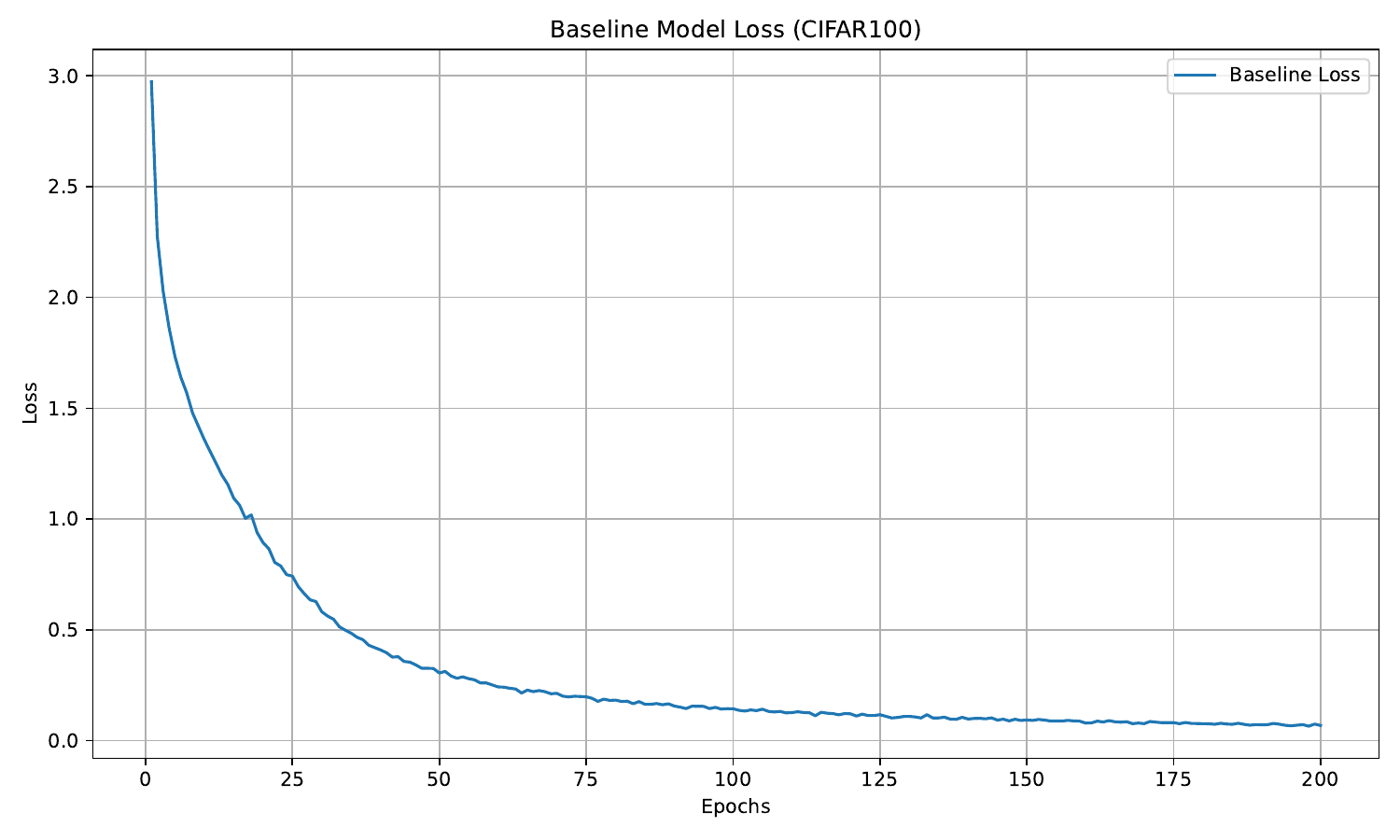}
    \caption{Baseline Model Loss for CIFAR100 with data augmentation. This figure illustrates the training loss for the baseline model over 200 epochs.}
    \label{fig:baseline_loss}
\end{figure}

\begin{figure}[h!]
    \centering
    \includegraphics[width=\linewidth]{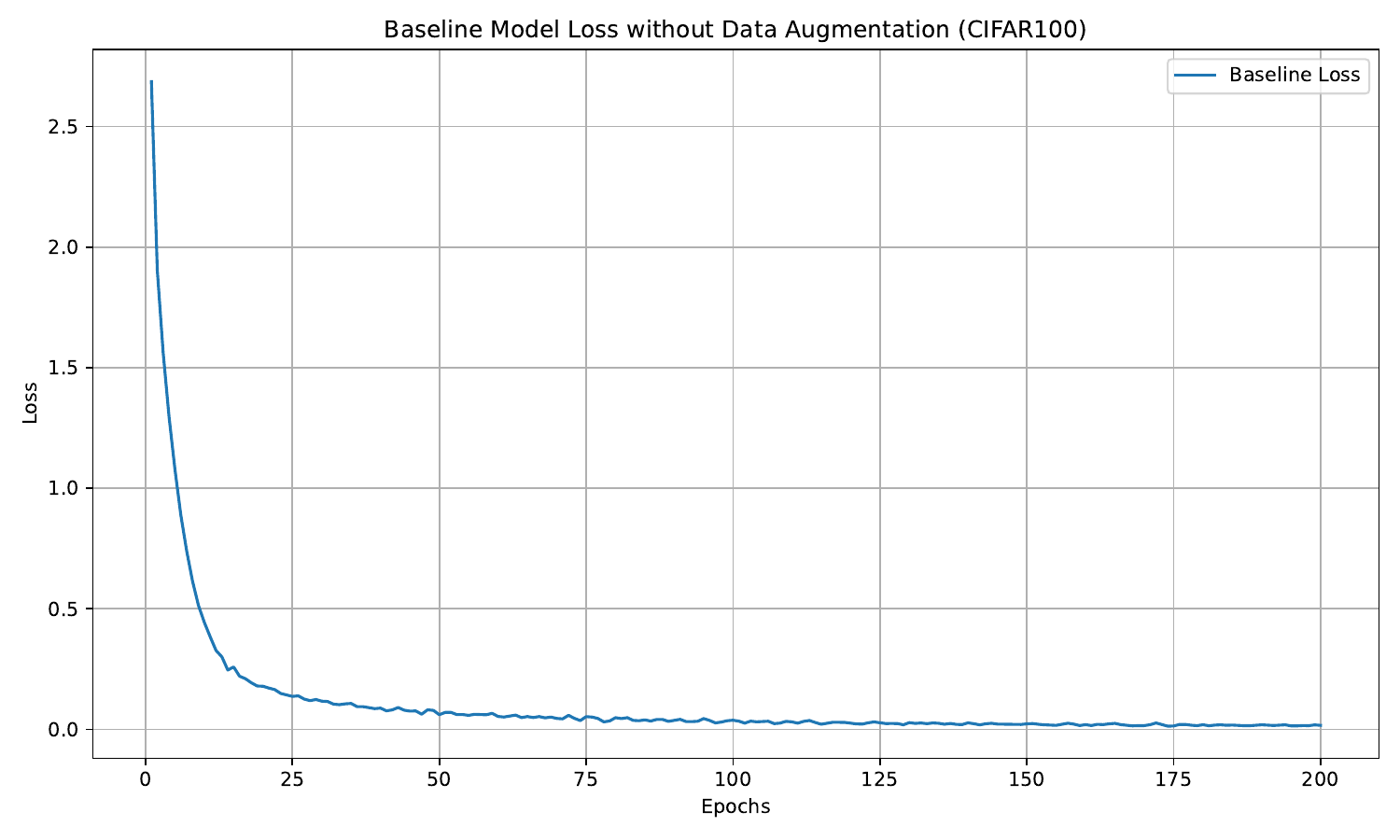}
    \caption{Baseline Model Loss Without Data Augmentation for CIFAR100. This figure shows the impact of excluding data augmentation during training.}
    \label{fig:baseline_no_aug}
\end{figure}

\begin{figure}[h!]
    \centering
    \includegraphics[width=\linewidth]{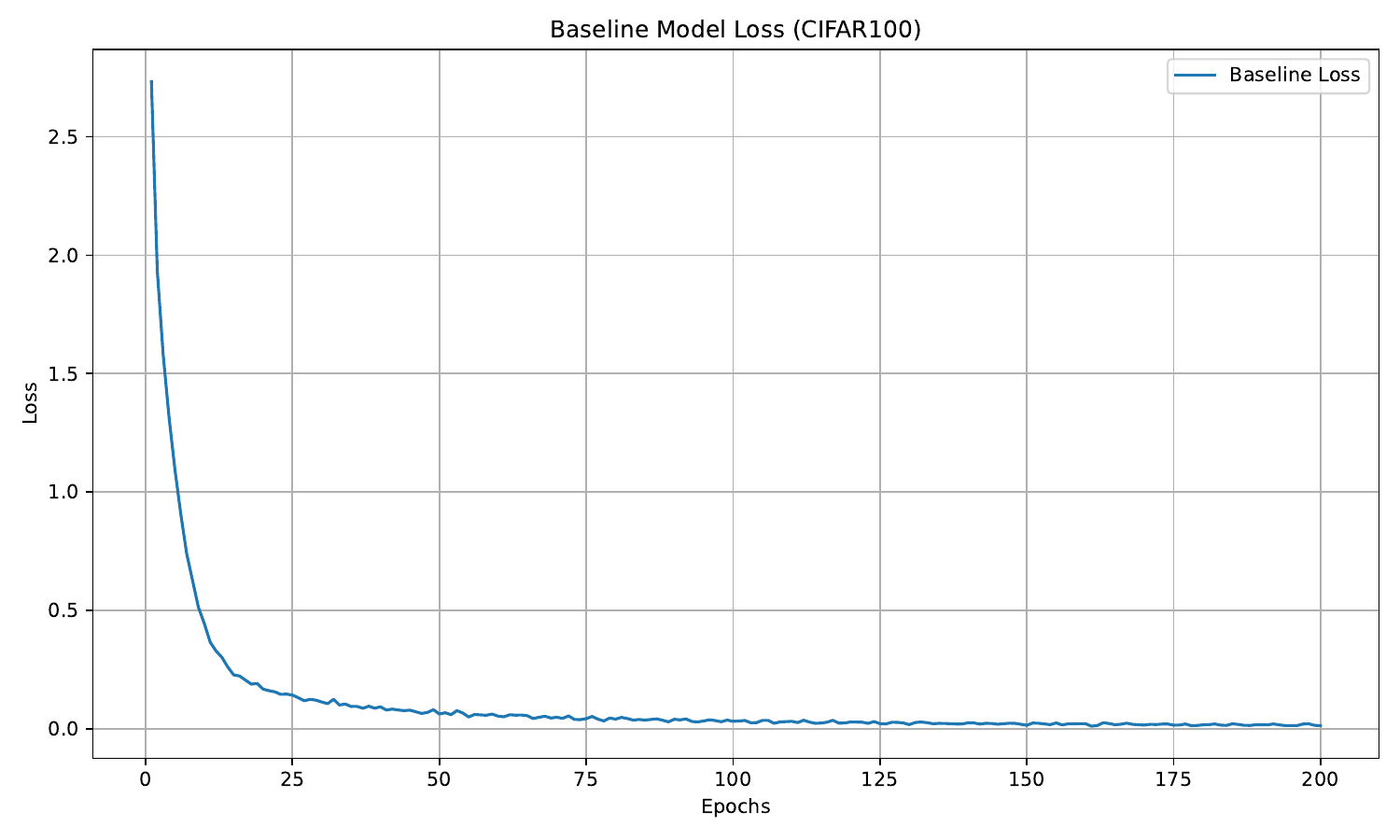}
    \caption{Baseline Model Loss with Margin SCL for CIFAR100. This figure highlights the baseline loss when using margin SCL.}
    \label{fig:baseline_margin_scl}
\end{figure}

\begin{figure}[h!]
    \centering
    \includegraphics[width=\linewidth]{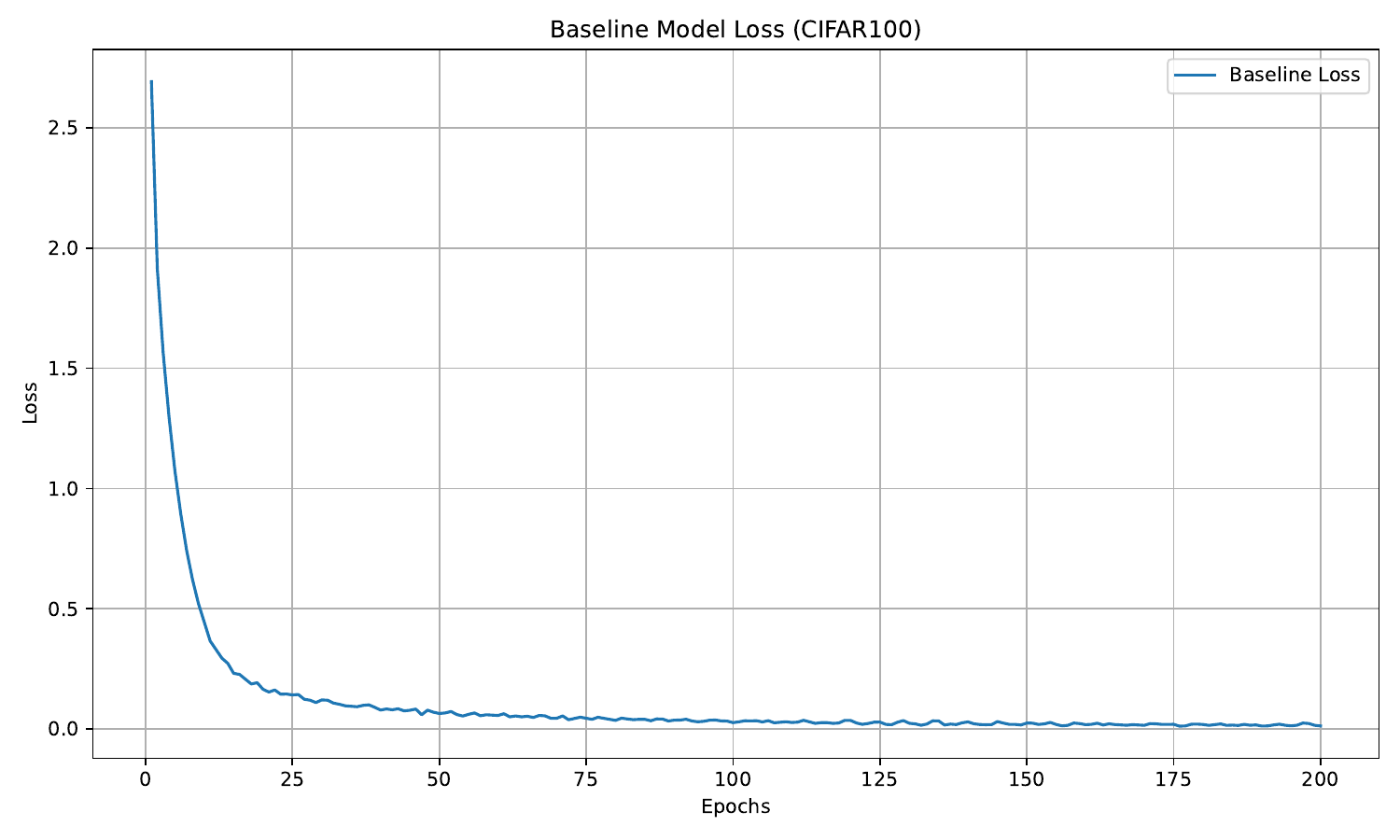}
    \caption{Baseline Model Loss with Refined SCL for CIFAR100. This figure illustrates the baseline model loss with refinements using the SCL approach.}
    \label{fig:baseline_refined_scl}
\end{figure}

\textbf{Baseline Model Loss with Data Augmentation:} The baseline ResNet-18 model was trained using cross-entropy loss with standard data augmentation techniques, including random cropping and horizontal flipping. This configuration represents a widely used benchmark for evaluating model performance on CIFAR-100.

\textbf{Baseline Model Loss Without Data Augmentation:} To isolate the impact of data augmentation, the baseline model was trained without any augmentation techniques. This configuration allowed for a direct comparison of the model's performance and loss dynamics when solely relying on clean data.

\textbf{Baseline Model Loss with Margin SCL:} The baseline model was refined using the margin-based contrastive loss. This approach integrates explicit margin constraints into the feature space to stabilize decision boundaries and improve robustness.

\textbf{Baseline Model Loss with Refined SCL:} This model applies the supervised contrastive loss (SCL) to the baseline model, refining its feature space by aligning embeddings of the same class and separating embeddings of different classes. This configuration highlights the role of SCL in restructuring the feature space for enhanced robustness.

\subsubsection{Supervised Contrastive Learning (SCL) Models}

\begin{figure}[h!]
    \centering
    \includegraphics[width=\linewidth]{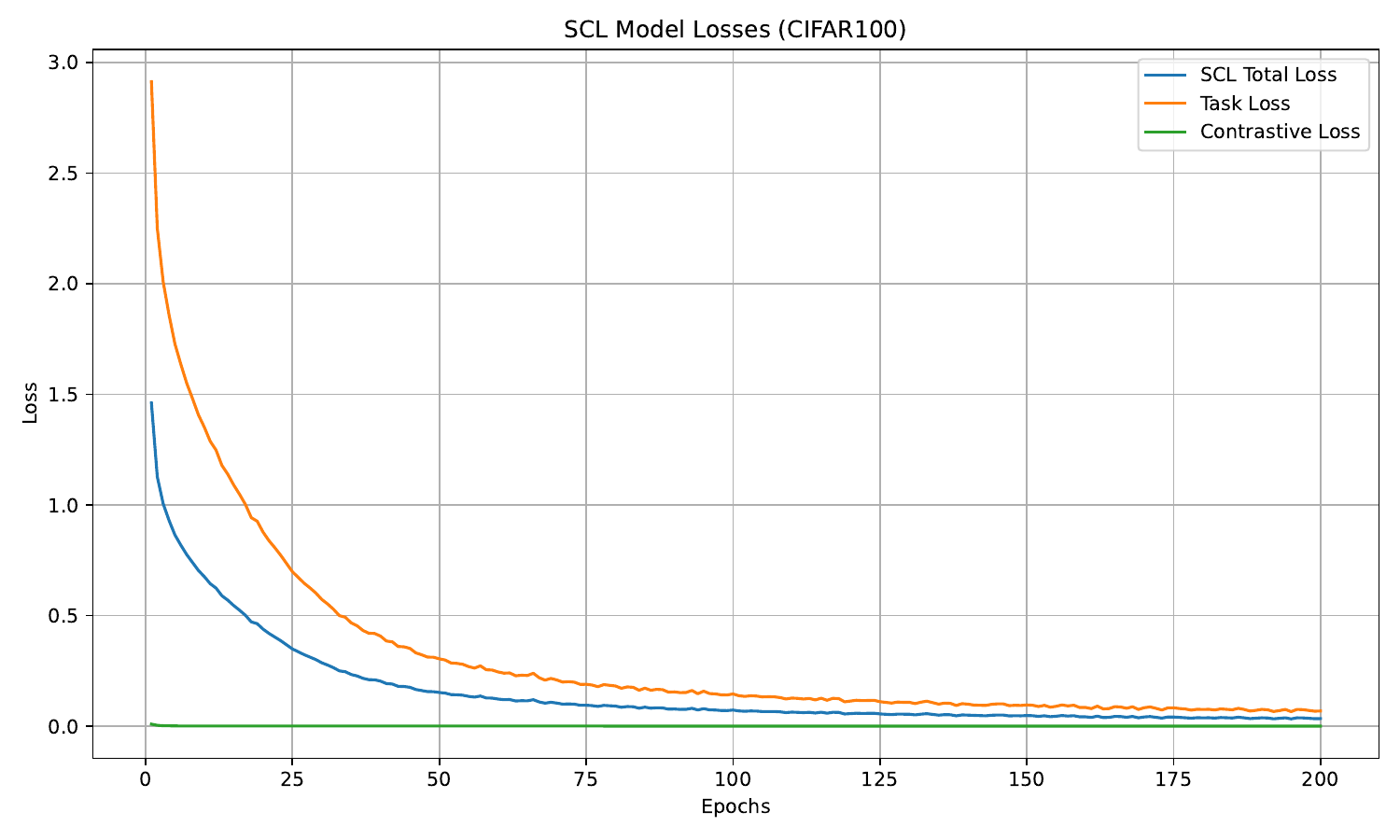}
    \caption{SCL Model Losses for CIFAR100 with data augmentation. Depicts the task loss, contrastive loss, and total loss for the SCL model over 200 epochs.}
    \label{fig:scl_losses}
\end{figure}

\begin{figure}[h!]
    \centering
    \includegraphics[width=\linewidth]{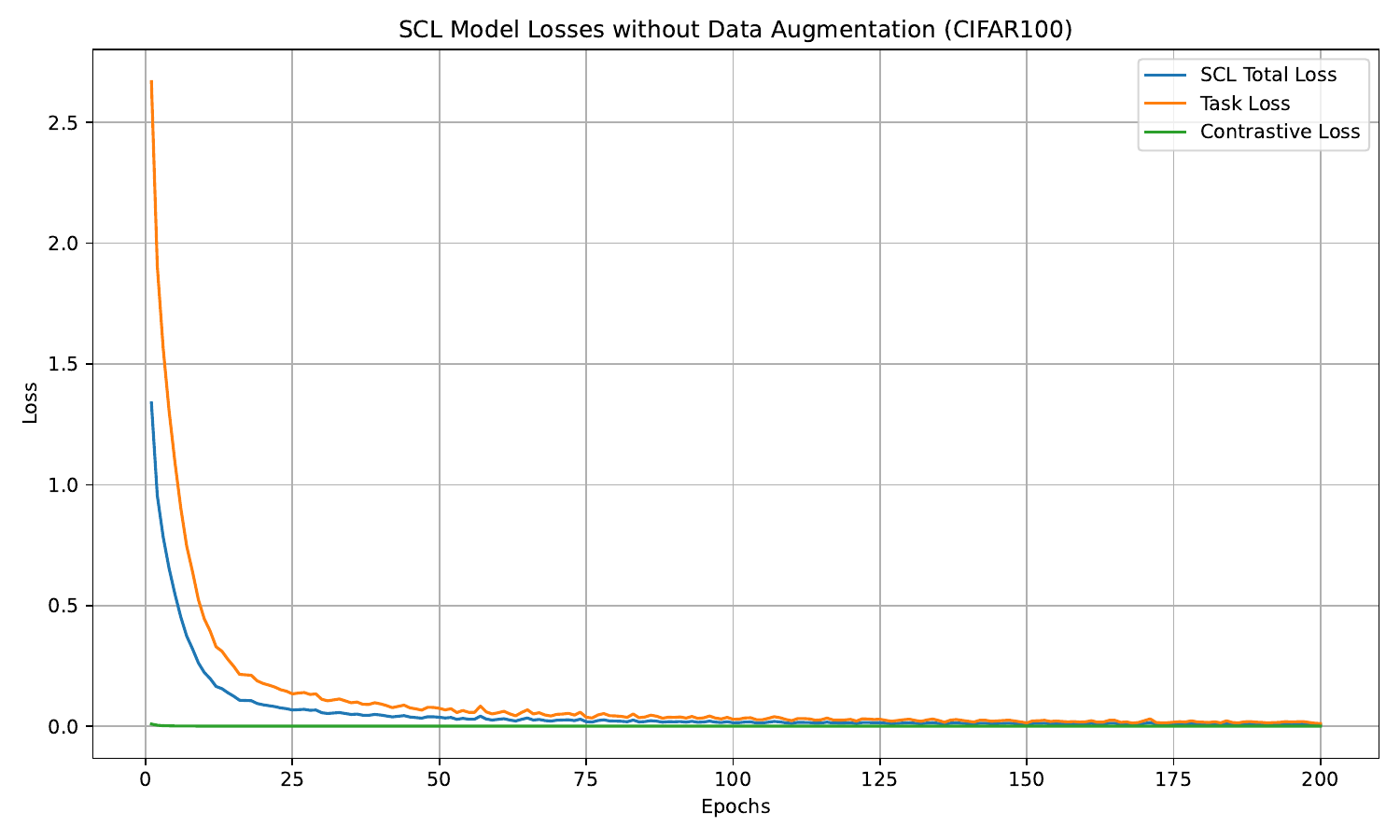}
    \caption{SCL Model Losses Without Data Augmentation for CIFAR100. Highlights how the model behaves without the use of augmentation.}
    \label{fig:scl_no_aug}
\end{figure}

\begin{figure}[h!]
    \centering
    \includegraphics[width=\linewidth]{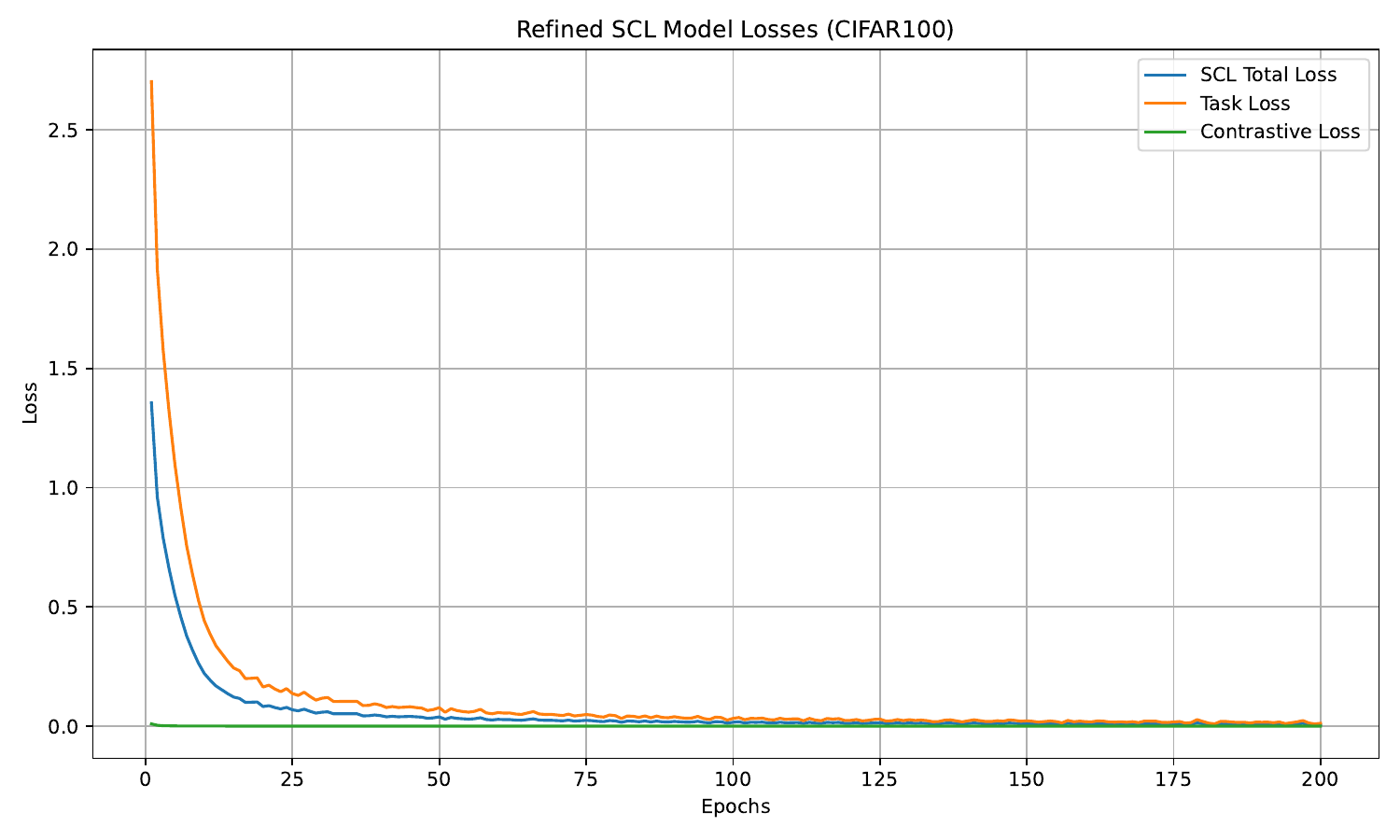}
    \caption{Refined SCL Model Losses for CIFAR100. This figure highlights the task loss, contrastive loss, and total loss for the refined SCL model over 200 epochs.}
    \label{fig:refined_scl_losses}
\end{figure}

\begin{figure}[h!]
    \centering
    \includegraphics[width=\linewidth]{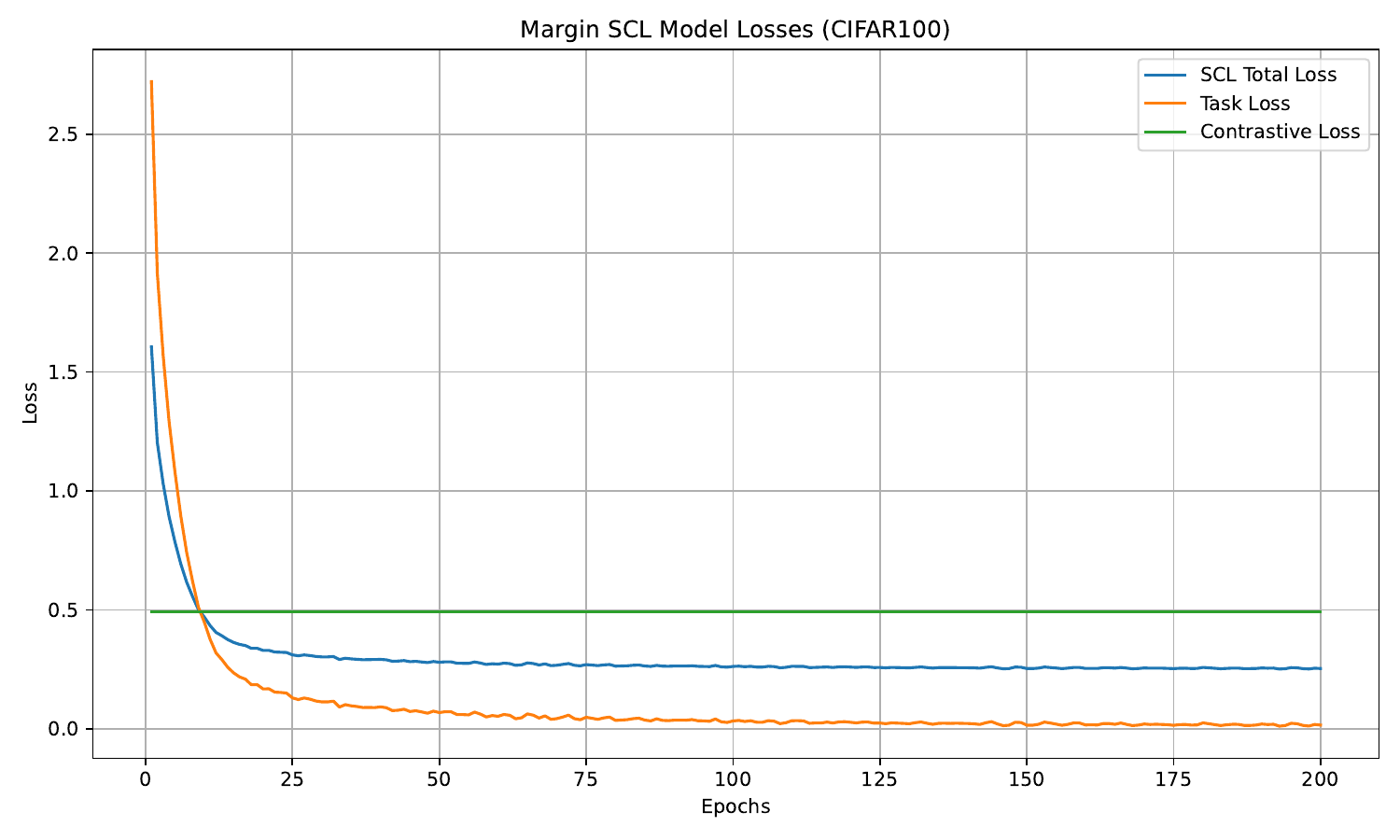}
    \caption{Margin SCL Model Losses for CIFAR100. This figure displays the task, contrastive, and total losses for the margin SCL model.}
    \label{fig:margin_scl_losses}
\end{figure}

\textbf{SCL Model Losses with Data Augmentation:} The supervised contrastive learning model was trained with data augmentation. The training dynamics included task loss, contrastive loss, and total loss over 200 epochs. This configuration emphasizes how data augmentation supports feature space clustering and separation during contrastive learning.

\textbf{SCL Model Losses Without Data Augmentation:} To examine the behavior of the SCL model without augmentation, this configuration focused on the same loss components (task loss, contrastive loss, and total loss) but trained exclusively on clean data. This comparison illustrates the dependency of contrastive learning on data augmentation.

\subsubsection{Refined SCL Models}

\textbf{Refined SCL Model Losses for CIFAR-100:} This model builds on the SCL approach by further optimizing the feature space through supervised contrastive loss combined with cross-entropy. The task loss, contrastive loss, and total loss were analyzed over 200 epochs to assess the impact of refinement on the model's robustness.

\textbf{Margin SCL Model Losses for CIFAR-100:} The margin-based contrastive learning model integrates explicit margin constraints with SCL. This configuration tracks the task loss, contrastive loss, and total loss, demonstrating the stability and robustness introduced by margin constraints in the feature space.

\subsection{FGSM Adversarial Accuracy}

To evaluate the robustness of the proposed models under adversarial conditions, we conducted experiments using the Fast Gradient Sign Method (FGSM) attack. The FGSM attack generates adversarial samples by perturbing the input data along the gradient direction of the loss function, controlled by a perturbation parameter $\epsilon$. This subsection provides a comparative analysis of FGSM accuracy for three different configurations: Baseline Model vs. SCL Model, Baseline Model vs. Refined SCL Model, and Baseline Model vs. Margin SCL Model.

\subsubsection{Baseline Model vs. SCL Model (Without Data Augmentation)}

\begin{figure}[h!]
    \centering
    \includegraphics[width=\linewidth]{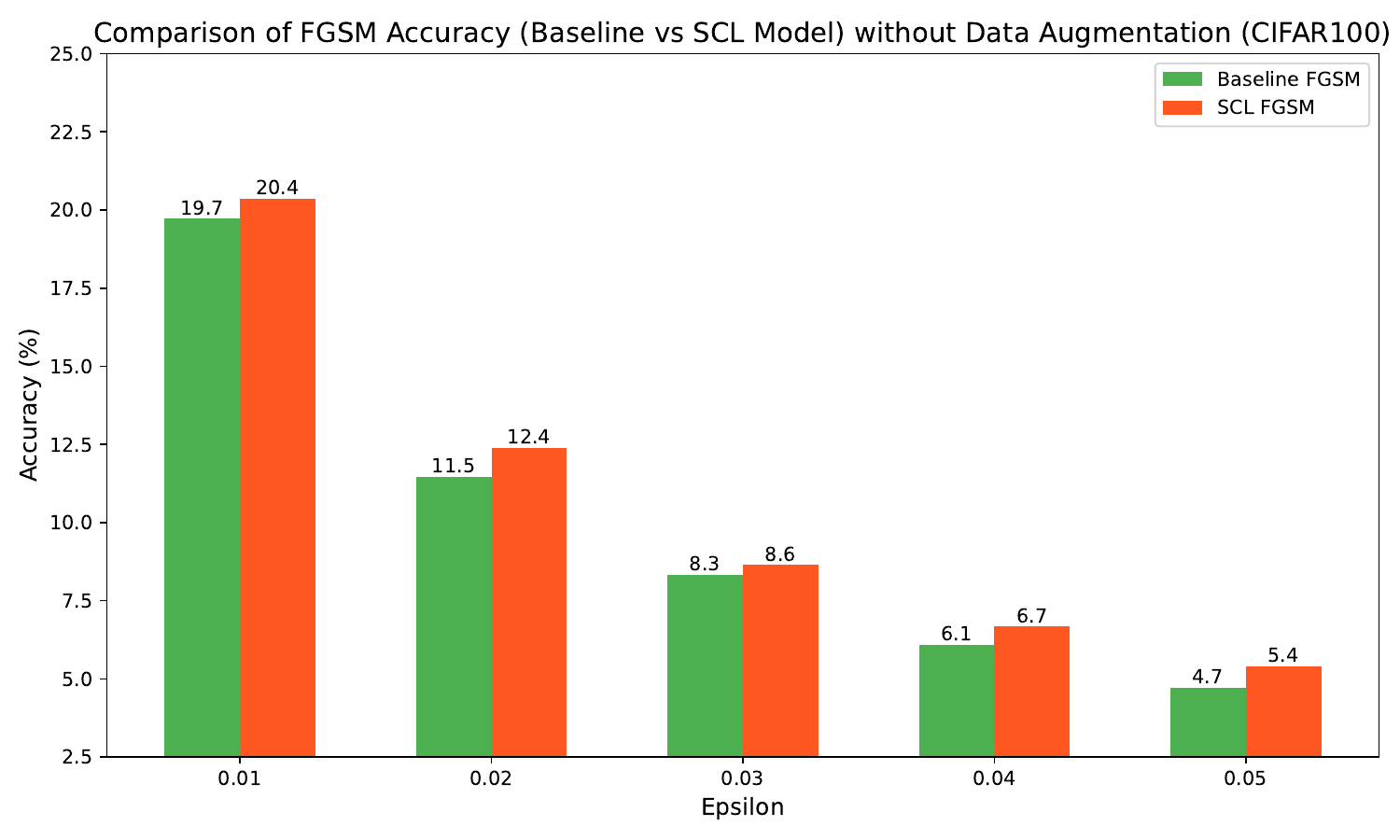}
    \caption{Comparison of FGSM Accuracy (Baseline vs SCL Model) Without Data Augmentation for CIFAR100.}
    \label{fig:fgsm_comp}
\end{figure}

Figure \ref{fig:fgsm_comp} compares the FGSM accuracy of the baseline model and the SCL-refined model trained without data augmentation. As $\epsilon$ increases, both models experience a decrease in accuracy due to larger perturbations. However, the SCL model consistently outperforms the baseline across all $\epsilon$ values. For example, at $\epsilon = 0.01$, the baseline achieves an accuracy of 19.7\%, while the SCL model achieves 20.4\%. This improvement demonstrates the effectiveness of supervised contrastive learning in creating more robust feature representations, even in the absence of data augmentation.

\subsubsection{Baseline Model vs. Refined SCL Model}

\begin{figure}[h!]
    \centering
    \includegraphics[width=\linewidth]{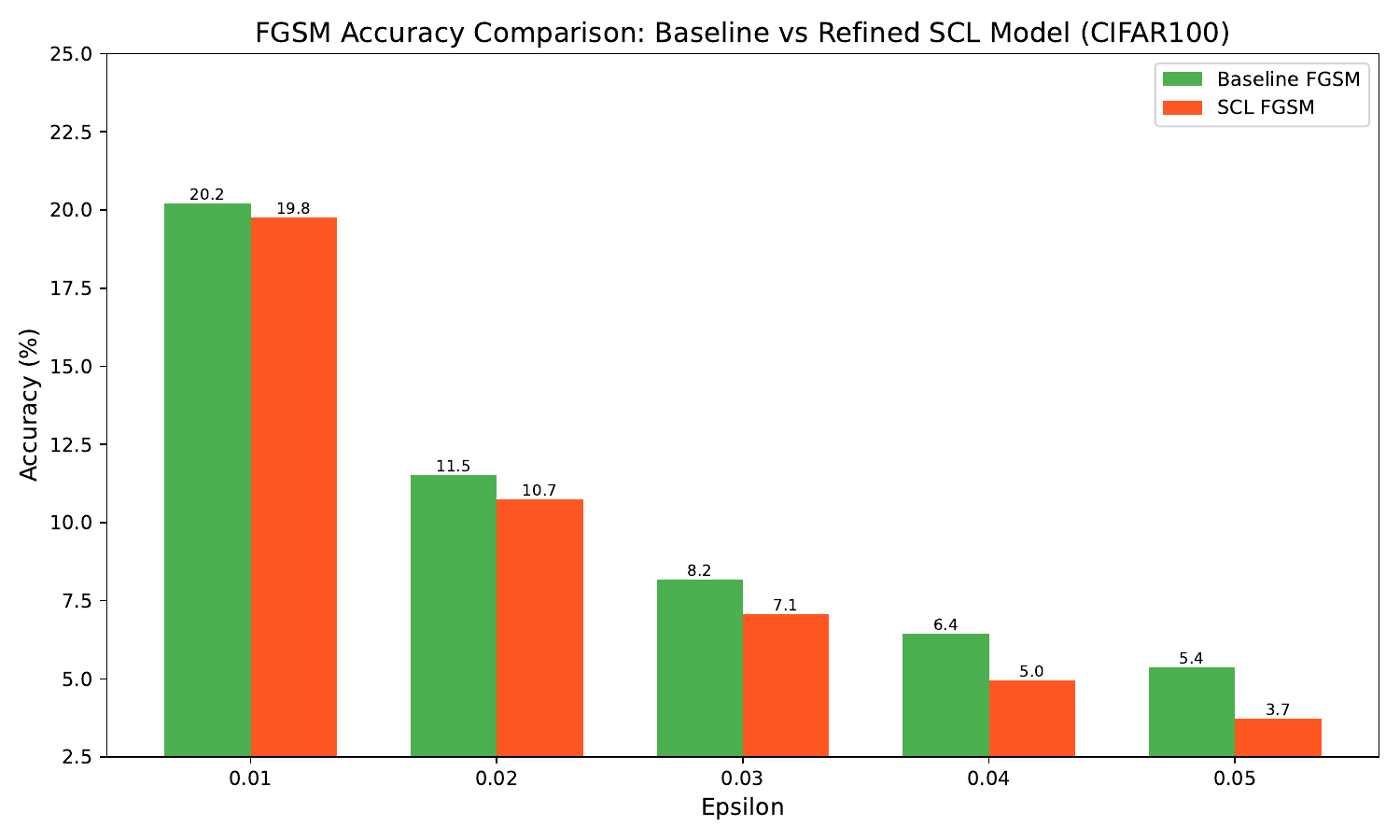}
    \caption{FGSM Accuracy Comparison: Baseline vs Refined SCL Model for CIFAR100.}
    \label{fig:fgsm_refined}
\end{figure}

Figure \ref{fig:fgsm_refined} compares the FGSM accuracy of the baseline model with the refined SCL model. The refined model is trained in two stages: first, the baseline model is trained using cross-entropy loss, and then it is retrained using supervised contrastive loss combined with the task-specific loss. Despite the additional refinement process, the refined SCL model does not consistently outperform the baseline under FGSM attacks. For instance, at $\epsilon = 0.01$, the baseline model achieves an accuracy of 20.2\%, which is slightly higher than the refined SCL model's 19.8\%. As $\epsilon$ increases, the refined model continues to perform worse than the baseline, achieving 7.1\% accuracy compared to the baseline's 8.2\% at $\epsilon = 0.03$.

This decline in performance suggests that retraining the model with contrastive loss after baseline training may not effectively enhance adversarial robustness. The potential reasons for this limitation include overfitting during the refinement stage or insufficient alignment of the learned feature representations with the contrastive loss objectives. These findings indicate the need for alternative approaches, such as joint training of baseline and contrastive objectives from the outset, to improve robustness.

\subsubsection{Baseline Model vs. Margin SCL Model}

\begin{figure}[h!]
    \centering
    \includegraphics[width=\linewidth]{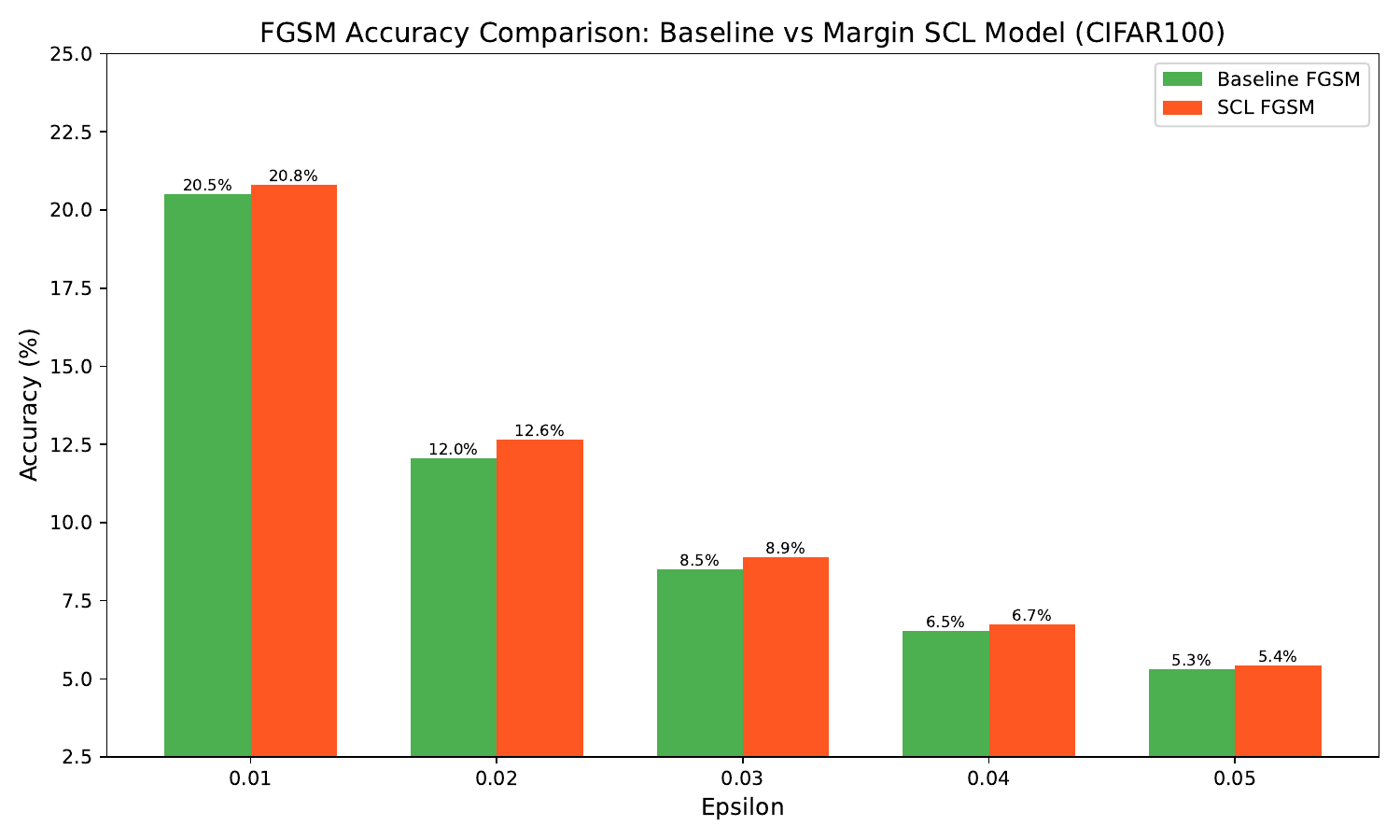}
    \caption{FGSM Accuracy Comparison: Baseline vs Margin SCL Model for CIFAR100.}
    \label{fig:fgsm_margin}
\end{figure}

Figure \ref{fig:fgsm_margin} compares the FGSM accuracy of the baseline model and the margin SCL model. The margin SCL model incorporates margin-based contrastive loss, explicitly enforcing intra-class compactness and inter-class separation. Across all $\epsilon$ values, the margin SCL model achieves slightly higher accuracy compared to the baseline. For instance, at $\epsilon = 0.02$, the margin SCL model achieves 12.6\% accuracy, outperforming the baseline at 12.0\%. This demonstrates that the margin-based constraints stabilize decision boundaries, resulting in enhanced robustness to FGSM attacks.

The comparative analysis of FGSM accuracy highlights the following:
\begin{itemize}
    \item Supervised contrastive learning (SCL) improves robustness compared to the baseline when trained directly with contrastive loss, particularly in scenarios without data augmentation.
    \item Refining the baseline model by retraining with contrastive loss after initial cross-entropy training does not significantly improve robustness and, in some cases, results in reduced performance under adversarial attacks. This indicates the need for better integration of contrastive learning objectives during initial training.
    \item Margin-based constraints consistently provide additional stability and robustness, outperforming both the baseline and refined SCL models across all perturbation levels.
\end{itemize}

These results validate the effectiveness of SCL and margin-based loss for improving adversarial robustness but also highlight the limitations of post-hoc refinement strategies.
The results demonstrate the effectiveness of the proposed framework in improving adversarial robustness. Supervised contrastive learning restructures the feature space by aligning embeddings of positive samples and separating embeddings of negatives, leading to better clustering and separation. The margin-based constraints further stabilize decision boundaries, creating a robust feature space that resists adversarial attacks.

\section{Conclusion}

The proposed framework combines supervised contrastive learning (SCL) with margin-based contrastive loss to enhance the adversarial robustness of convolutional neural networks (CNNs). By aligning embeddings within the same class and separating those from different classes, SCL creates a structured feature space, while margin-based constraints stabilize decision boundaries through explicit intra-class compactness and inter-class separation. Experiments on CIFAR-100 demonstrate that the proposed approach achieves performance increase in adversarial accuracy under FGSM attacks, without compromising clean data performance. These results establish the proposed framework as an efficient solution for robust deep learning, with potential for broader applications and integration with other adversarial defense mechanisms.

\end{document}